\documentclass{article}
\usepackage{spconf,amsmath,graphicx}
\usepackage{url}
\usepackage{booktabs}
\usepackage{amsmath}
\usepackage{amssymb}
\usepackage{amsfonts}
\usepackage{amsthm}
\usepackage{color}
\usepackage{amsmath,amsfonts}
\usepackage{amsthm}
\usepackage{amssymb,amsopn}
\usepackage{bm} 

\newcommand{\vct}[1]{\boldsymbol{#1}} 
\newcommand{\mat}[1]{\boldsymbol{#1}} 

\newcommand{\eat}[1]{}

\newcommand{\ie}{\textit{i}.\textit{e}.}
\newcommand{\eg}{\textit{e}.\textit{g}.}

\title{RoutingGAN: Routing Age Progression and Regression with Disentangled Learning}
\name{Zhizhong Huang$^{1}$ \qquad Junping Zhang$^{1}$ \qquad Hongming Shan$^{2,*}$\thanks{Corresponding author}\thanks{© 2021 IEEE.  Personal use of this material is permitted. Permission from IEEE must be obtained for all other uses, in any current or future media, including reprinting/republishing this material for advertising or promotional purposes, creating new collective works, for resale or redistribution to servers or lists, or reuse of any copyrighted component of this work in other works.}}

\address{$^{1}$Shanghai Key Lab of Intelligent Information Processing, School of Computer Science\\
$^{2}$Institute of Science and Technology for Brain-inspired Intelligence \\Fudan University, Shanghai 200433, China
}

\begin{document}
\graphicspath{{figs/}}
%
\maketitle
\begin{abstract}
Although impressive results have been achieved for age progression and regression, there remain two major issues in generative adversarial networks (GANs)-based methods: 1) conditional GANs (cGANs)-based methods can learn various effects between any two age groups in a single model, but are insufficient to characterize some specific patterns due to completely shared convolutions filters; and 2) GANs-based methods can, by utilizing several models to learn effects independently, learn some specific patterns, however, they are cumbersome and require age label in advance. To address these deficiencies and have the best of both worlds, this paper introduces a dropout-like method based on GAN~(RoutingGAN) to route different effects in a high-level semantic feature space. Specifically, we first disentangle the age-invariant features from the input face, and then gradually add the effects to the features by residual routers that assign the convolution filters to different age groups by dropping out the outputs of others. As a result, the proposed RoutingGAN can simultaneously learn various effects in a single model, with convolution filters being shared in part to learn some specific effects. Experimental results on two benchmarked datasets demonstrate superior performance over existing methods both qualitatively and quantitatively.
\end{abstract}

%
\section{Introduction}
Age progression and regression, also known as face aging and rejuvenation, aim at rendering a given face to predict its appearance at different ages with natural effects while preserving personal identity. It has broad applications ranging from digital entertainment to information forensics and security including face age editing and cross-age face verification~\cite{wu2012age}. Despite the appealing practical value, age progression and regression remain challenging due to the lack of labeled age data of the same subject and the intrinsic complexity in rendering natural effects with the identity consistently preserved.

\begin{figure*}[t]
\centering
\includegraphics[width=1.\linewidth]{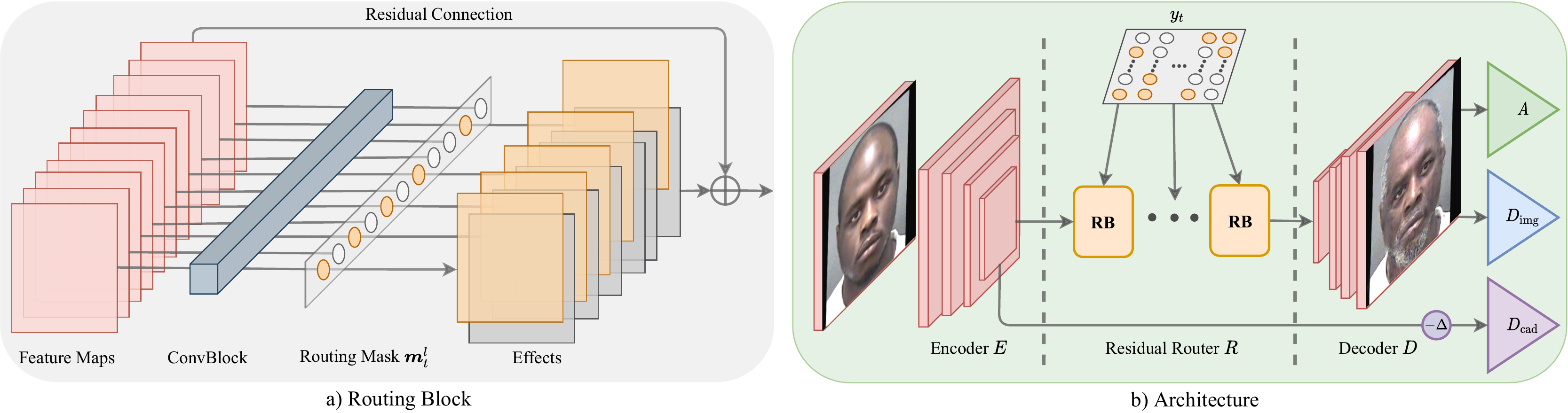}
\vspace{-7mm}
\caption{The proposed RoutingGAN. a) A basic routing block in proposed residual router uses a mask to assign filters to different age groups by dropping out some feature maps along channels. b) The overall architecture of the proposed method, where RB denotes routing block.}
\label{framework}
\end{figure*}

In recent years, generative adversarial networks~(GANs)~\cite{goodfellow2014generative}, especially conditional GANs~(cGANs), have been used to tackle this issue by training with unpaired data. These approaches have shown more promising results than traditional methods~\cite{wang2016recurrent} in terms of three underlying requirements for age progression and regression---image quality, identity preservation, and age accuracy. The resultant methods can be roughly summarized into two categories: cGANs- and GANs-based methods. The key difference between these two different kinds of methods turns out that cGANs-based methods~\cite{zhang2017age,wang2018face,li2018global,zhu2020look} are more flexible than GANs-based methods but GANs-based methods~\cite{yang2018learning,liu2019attribute,huang2020pfa} can produce better results in age progression and regression. Therefore, it is desirable to have a single model that could learn various aging effects independently, at least in part.

To this end, we propose a dropout-like method based on generative adversarial network~(RoutingGAN) that assigns the filters to different age groups, routing the different effects in a high-level semantic feature space. Specifically, the generator of our RoutingGAN consists of three main components: 1) an encoder that disentangles the age-invariant features from the input face; 2) a residual router that routes and gradually applies the effects to the disentangled features by dropping out the output of specific filters; and 3) a decoder that synthesizes the faces from the transformed features. Different from original Dropout~\cite{hinton2012improving} that randomly omits features with a fixed probability during training, the convolution filters in the residual router are assigned at network initialization but fixed during training and testing with a ratio $\sigma$; \eg, $\sigma=0.5$ indicates that half filters are kept for each age mapping. Therefore, filters in the residual router can focus on only one single transformation patterns independently to learn some unique effects or be shared across different age groups simultaneously to learn the general ones.

The contributions of this paper are summarized as follows. 1) We propose a novel approach for age progression and regression that enables existing cGANs-based methods to learn various effects independently, at least in part, by residual routers that assign convolution filters to different age groups. As a result, our method can significantly improve performance.  Note that our work is orthogonal to the existing research direction~\cite{zhang2017age,wang2018face,li2018global,zhu2020look,li2019age} and can be easily integrated into current methods by only modifying the architecture of the generator while preserving their own advantages. 2) Extensive experiments on two benchmarked datasets demonstrate the effectiveness and robustness of the proposed method in rendering accurate effects while preserving identity through both qualitative and quantitative comparisons.
\section{Methodology}

In this section, we first describe the disentangled learning that disentangles the age-invariant features from input faces, then introduce the proposed residual router that empowers the existing cGANs-based methods with multiple models, and finally detail the whole framework, followed by the summary of the loss functions. Following~\cite{li2018global,zhu2020look,yang2018learning,liu2019attribute}, we divide all ages into $N=4$ non-overlapping age groups; \ie, $30-$, $31-40$, $41-50$, and $51+$.

\subsection{Disentangled Learning}

Given an input face image $\mat{X}_s \in\mathbb{R}^{h\times w\times 3}$ with age group $y_s$, the encoder $E$ extracts features $E(\mat{X}_s)$, which are supposed to be invariant from ages since the effects are gradually applied by the residual router $R$. In other words, if $E(\mat{X}_s)$ contains the age information, the gap between features at different ages leads to strong ghosts in the generated faces. Therefore, a cross-age domain discriminator $D_{\mathrm{cad}}$ with a gradient reversal layer~\cite{ganin2016domain} is incorporated with the architecture. 

Specifically, $D_\mathrm{cad}$ is updated to distinguish the age groups of the $E(\mat{X}_s)$ while the gradient reversal layer has the encoder $E$ updated towards the opposite direction of the gradient. Consequently, $E$ and $D_\mathrm{cad}$ are competing against each other in an adversarial manner during training, which enforces $E$ to extract age-invariant features. Furthermore, to reduce the computational cost, the encoder has 4 convolutional layers that have respectively 64, 128, 256, and 512 $4\times 4$ stride-2 convolution filters while $D_\mathrm{cad}$ has only two fully-connected layers with 512 and $N$ neurons, respectively. Formally, the loss function to optimize both $E$ and $D_\mathrm{cad}$ is defined as:
\begin{align}
    \mathcal{L}_{\mathrm{cad}}= \mathbb{E}_{\mat{X}_{s}}\left[\ell\left(D_\mathrm{cad}(E(\mat{X}_s)), y_{s}\right)\right],
\end{align}
where $\ell$ is the cross-entropy loss for age group classification.

\subsection{Residual Router}

In this subsection, we introduce our proposed dropout-like residual router. Dropout~\cite{hinton2012improving} has been successfully applied in numerous deep-learning tasks such as image classification~\cite{he2016deep} and many-task learning~\cite{strezoski2019many}. It deals with overfitting by randomly omitting some hidden units of the neural network with a fixed probability of $p$ so that every time the network optimizes a resulting sub-network based on a mini-batch of data during training and ensembles all sub-networks at the testing stage. In a sense, a huge number of different sub-networks reside in a single network with shared weights, which makes it possible to equip existing cGANs-based methods with multiple models for different patterns in a similar way. However, directly applying dropout to cGANs-based methods is not applicable since they have to render the input faces conditioned on the target ages. Therefore, we propose to achieve it in a similar way by assigning different filters to different age groups where some filters are shared across age groups and some are unique for some specific age group. 

Specifically, as shown in Fig.~\ref{framework}(a), we append a binary mask $\vct{m}^l_t \in \mathbb{R}^{1\times 1\times C}$ at the end of the residual output of the routing block, where $C$ is the number of convolution filters, $l$ is the index of the residual routing block, and $t$ is the index of the target age group. Furthermore, all of the unique binary masks form a binary tensor $\mat{M}\in \mathbb{R}^{L\times N\times C}$ for the residual router, where $L$ is the number of routing block. Here, we use a hyper-parameter $\delta$ to control the number of used filters for each age group at each routing block. For example, one group only uses half of the filters when $\sigma=0.5$. Note that $\mat{M}$ is randomly initialized at the beginning and fixed during training and testing. 

Consequently, the only difference from existing cGAN-based methods is that RoutingGAN needs to select the corresponding binary mask of the target age $t$ during age progression and regression. Therefore, RoutingGAN preserves the flexibility of cGANs-based methods while the computational cost can be significantly reduced compared to GANs-based methods as it puts all sub-models in one. Besides, the shared filters can focus on the general effects while the specific effects can be learned by the unique filters, making our method produce better results than cGANs-based methods.

\subsection{Framework} 

In this subsection, we detail the remaining main components of RoutingGAN, which is shown in Fig.~\ref{framework}(b).

\noindent\textbf{Decoder} After encoding and transforming by the encoder $E$ and residual router $R$, respectively, the features are fed into the decoder $D$. Specifically, the decoder has 4 deconvolutional layers followed by a tanh activation function, whose filter sizes and the numbers of filters are the same as the encoder, but in reverse order. Given a target age group $y_t$, the process of rendering $\mat{X}_s$ can be formulated as:
\begin{align}
    \mat{\widehat{X}}_t = D\Big(R\bigl(E(\mat{X}_s), \vct{y}_{t}\bigr)\Big).
\end{align}

\noindent\textbf{Image Discriminator} We adopt the PatchDiscriminator from~\cite{zhu2017unpaired} as our discriminator $D_{\mathrm{img}}$ to distinguish the generated faces from the real ones. It has a series of 6 convolutional layers with an increasing number of $4\times 4$ filters, each of which, except the first and last layer, is followed by a spectral normalization layer~\cite{miyato2018spectral} and a LeakyReLU activation with a slope of 0.2 for negative input. Besides, $\mat{C}_t$ is concatenated with the feature maps of the first convolutional layer in order to align conditions with the generated images. Here, we employ the least-squares GANs~\cite{mao2017least} to stabilize our training process. Specifically, it adopts the least-squares loss function rather than the negative log-likelihood to force the generator to generate samples toward the decision boundary. The adversarial loss to optimize the encoder, residual router, and decoder is thus defined as:
\begin{align}
    \mathcal{L}_{\mathrm{adv}}=\frac{1}{2} \mathbb{E}_{\mat{X}_{s}}\Big[D_\mathrm{img}\big([\mat{\widehat{X}}_t; \mat{C}_t]\big)-1\Big]^{2}.
\end{align}

\begin{figure}[!t]
    \centering
    \includegraphics[width=0.8\linewidth]{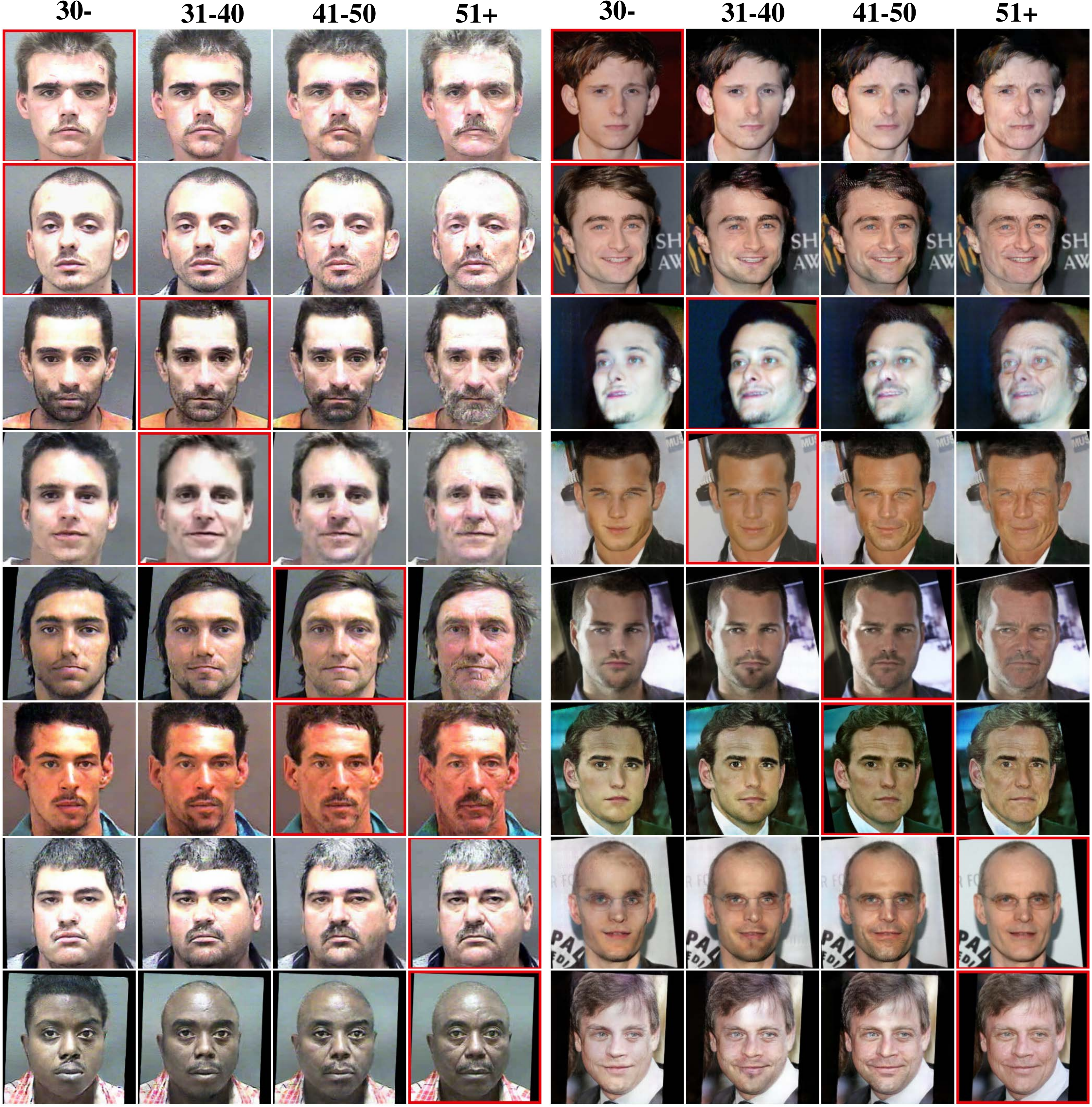}
    \caption{Sample results of both face aging and rejuvenation by applying our RoutingGAN to MORPH~(left) and CACD~(right). Red boxes indicate input faces.}
    \label{results}
\end{figure}

\noindent\textbf{Age Classifier} Following~\cite{wang2018face,li2018global,zhu2020look,li2019age}, we employ the pre-trained age classifier $A$ to further improve the age accuracy. We train a ResNet50 and a VGG16, then fix and ensemble them as $A$ for deep age supervision. Therefore, the age classification loss between the estimated age of generated faces and the target age $y_t$ is written as:
\begin{align}
    \mathcal{L}_{\mathrm{age}}=\mathbb{E}_{\mat{X}_{s}}\left[\ell\bigl(A(\mat{\widehat{X}}_t), y_{t}\bigr)\right].
\end{align}

\begin{figure*}[t]
    \centering
    \includegraphics[width=0.72\linewidth]{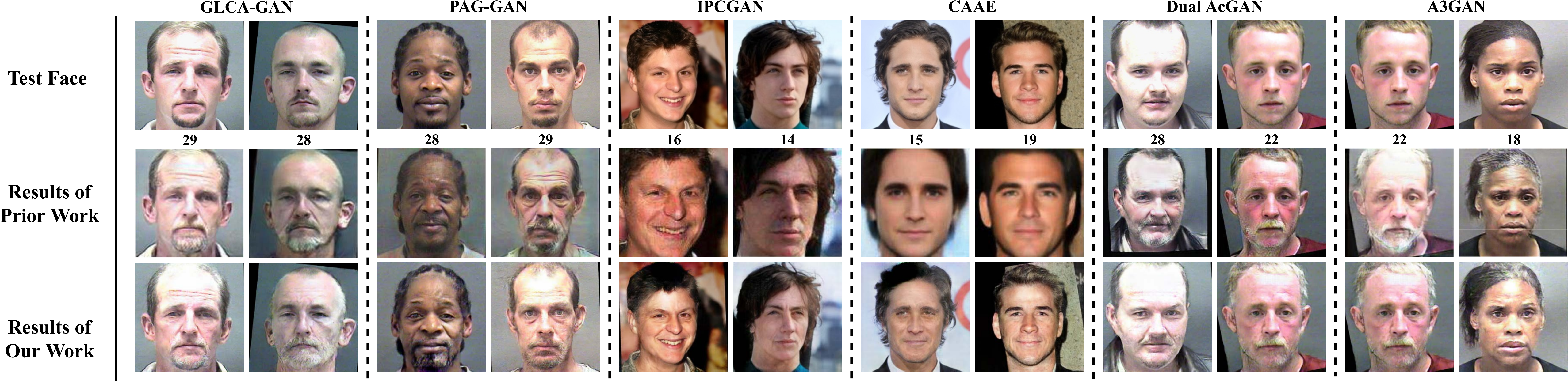}
    \caption{Performance comparison with prior work on the MORPH and CACD datasets. The three rows are the input young faces with their real age labels underneath, the results of prior work, and our results in the same age groups~($51+$), respectively. 
    }
    \label{quality_comparsion}
\end{figure*}

\noindent\textbf{Loss Functions}
To maintain the identity consistency and close the input-output gap, \eg, the color aberration, we adopt two losses between the input face and generated one, including a pixel-wise loss, and a feature-level loss. The two losses are defined as follows:
\begin{align}
    \mathcal{L}_{\mathrm{pix}} &= \mathbb{E}_{\mat{X}_{s}}\Big\|\mat{\widehat{X}}_t-\mat{X}_{s}\Big\|^2_F, \\
      \mathcal{L}_{\mathrm{fea}}&=\mathbb{E}_{\mat{X}_{s}}\Big\|\phi\Bigl(\mat{\widehat{X}}_t\Bigr)-\phi(\mat{X}_{s})\Big\|^2_F.
\end{align}
where $\phi$ denotes the activation output of the $10$th convolutional layer from the VGG-Face descriptor~\cite{parkhi2015deep} and $\|\cdot\|_F$ represents the Frobenius norm.

Finally, the overall loss function to optimize the encoder $E$, residual router $R$, decoder $D$, and cross-age domain discriminator $D_{\mathrm{cad}}$ is expressed as:
\begin{align}
    \mathcal{L}=\lambda_{\mathrm{adv}} \mathcal{L}_{\mathrm{adv}} + \lambda_{\mathrm{age}} \mathcal{L}_{\mathrm{age}} + \lambda_{\mathrm{fea}} \mathcal{L}_{\mathrm{fea}} + \lambda_{\mathrm{pix}} \mathcal{L}_{\mathrm{pix}}.
\end{align}
where $\lambda_{\mathrm{*}}$ controls the balance between different loss terms. Besides, the loss to optimize $D_\mathrm{img}$ is defined as follows:

\begin{footnotesize}
\begin{align}
    \mathcal{L}_{D_{\mathrm{img}}} =& \frac{1}{2} \mathbb{E}_{\mat{X}_s} \Bigl[ \Big(D_{\mathrm{img}}\big([\mat{X}_s;\mat{C}_s]\big)-1\Big)^{2} + D_{\mathrm{img}}\big([\mat{\widehat{X}}_t;\mat{C}_t]\big)^{2}\Bigr].
\end{align}
\end{footnotesize}

\section{EXPERIMENTS}

\subsection{Implementation Details}

We conducted experiments on two benchmarked age datasets: MORPH~\cite{ricanek2006morph} and CACD~\cite{chen2015face}. After aligning and cropping, we got a total of 159,585 and 55,139 images for the two datasets with a size of $256\times 256$. For each dataset, we randomly selected 80\% images for training and the remaining for testing, without identities overlapping, and all images are normalized into $[-1, 1]$. There are eight routing blocks in the residual router and $\sigma$ was set to $0.5$. We adopted Adam optimization method with a fixed learning rate of $10^{-4}$ for all modules, and the hyperparameters in loss functions shared across datasets were empirically set as follows: $\lambda_{\mathrm{adv}}$ was $75$, $\lambda_{\mathrm{pix}}$ was $5\times 10^{-4}$, $\lambda_{\mathrm{fea}}$ was $10^{-3}$, $\lambda_{\mathrm{age}}$ was $20$ and $\lambda_{\mathrm{cad}}$ was $10$. We trained all models with a mini-batch of size 32 on four 2080Ti
GPUs and 100,000 iterations.

\begin{table}[t]
    \centering
    \caption{Quantitative comparison in terms of age estimation error (AEE) and face verification rate (VR) on two datasets. Due to the limited space, we only report the mean value computed over all age mappings.
    }
    \label{tab:performance}
    \begin{tabular*}{\columnwidth}{@{\extracolsep{\fill}}lrrrrr}
    \toprule
    & \multicolumn{2}{c}{MORPH} &  & \multicolumn{2}{c}{CACD} \\ 
    \cmidrule{2-3}\cmidrule{5-6} 
    & AEE & VR~(\%) && AEE & VR~(\%) \\
    \midrule
    CAAE~\cite{zhang2017age}       & 10.34 & 34.83  && 5.16 & 3.59   \\
    IPCGAN~\cite{wang2018face}     & 1.74  & 99.86  && 8.11 & 99.19  \\
    Dual cGAN~\cite{song2018dual}  & 2.44  & 99.99  && 3.28 & 99.88  \\
    Dual AcGAN~\cite{li2019age} & 1.53  & \textbf{100.00} && 1.78 & \textbf{99.92}  \\
    RoutingGAN       & \textbf{1.16}  & \textbf{100.00} && \textbf{1.57}    & \textbf{99.92}     \\ 
    \bottomrule
    \end{tabular*}
    
\end{table}
\subsection{Qualitative Comparison}

Fig.~\ref{results} showcases some generated faces of both face aging and rejuvenation. Although input faces cover a wide range of the population in terms of race, gender, pose, makeup, and expression, the model successfully renders photo-realistic and diverse effects with natural details in the skin, muscles, wrinkles, etc. Besides, identity consistency is well preserved in all generated face images.

Fig.~\ref{quality_comparsion} presents the qualitative comparison between our method and other state-of-the-art methods including GLCA-GAN~\cite{li2018global}, PAG-GAN~\cite{yang2018learning}, IPCGAN~\cite{wang2018face}, CAAE~\cite{zhang2017age}, Dual AcGAN~\cite{li2019age}, and A3GAN~\cite{liu2019attribute}. RoutingGAN renders faces with more natural and detailed effects than cGANs-based methods~\cite{zhang2017age,wang2018face,li2018global}, and is better at suppressing ghosting artifacts and color distortion than~\cite{yang2018learning,liu2019attribute,li2019age}. Note that we directly refer to results from their published papers for a fair comparison; this strategy is widely used in the mainstream literature.

\subsection{Quantitative Comparison}

We further adopted two widely-used quantitative metrics to evaluate the performance of age progression and regression methods---age  accuracy and identity preservation. For age accuracy, we calculate the age estimation error (AEE) between estimated ages of real and fake face images of all age mappings. For identity preservation, we reported the verification rate (VR) of whether the input faces and the generated ones are the same persons. All metrics are conducted by invoking the publicly Face++ APIs~\cite{faceplusplus.com} for a fair comparison, and the threshold for face verification is set to $76.5$. RoutingGAN competes against the state-of-the-art methods such as CAAE, IPCGAN, Dual cGAN~\cite{song2018dual}, and Dual AcGAN since they are all cGANs-based methods and attempt to achieve age progression and regression in a single unified framework.

Table~\ref{tab:performance} shows that RoutingGAN outperforms other baseline methods by a large margin in both age translation accuracy and identity preservation. On one hand, CAAE over-smoothens face images with subtle changes, leading to high age estimation errors and low face verification rates while IPCGAN brings ghosts and unexpected changes with a single model for both age progression and regression. Besides, our method generates images with double resolution than CAAE and IPCGAN~($256\times 256$ vs. $128\times 128$). On the other hand, since Dual (A)cGAN consists of two separate models for face aging and rejuvenation, they are faced with the same problem as the GANs-based methods. In summary, RoutingGAN not only performs the best quantitatively but also preserves the essential flexibility of cGANs-based methods.

\section{conclusions}
In this paper, we proposed a dropout-like method for age progression and regression. It extends the existing cGAN-based methods to have residual routers that assign filters to different age groups. The experimental results demonstrate the effectiveness of our method both qualitatively and quantitatively on two benchmarked datasets.

\clearpage

\end{document}